\pdfoutput=1

\documentclass[11pt]{article}

\usepackage{acl}

\usepackage{times}
\usepackage{latexsym}

\usepackage[T1]{fontenc}

\usepackage[utf8]{inputenc}

\usepackage{microtype}

\usepackage{graphicx}
\usepackage{amsmath}
\usepackage{soul}
\usepackage{amsthm}
\usepackage{algpseudocode}
\usepackage{booktabs}
\usepackage[ruled,vlined,linesnumbered]{algorithm2e}
\usepackage[capitalize]{cleveref}
\usepackage{multirow,multicol}
\usepackage{colortbl}
\usepackage{natbib}

\usepackage{xcolor}

\DeclareMathOperator*{\KL}{\text{KL}}
\title{\textsc{DoubleMix}: Simple Interpolation-Based Data Augmentation \\ for Text Classification}

\author{Hui Chen$^\clubsuit$  Wei Han$^\clubsuit$  Diyi Yang$^\diamondsuit$  Soujanya Poria$^\clubsuit$\\
  $^\clubsuit$ Singapore University of Technology and Design\\
  $^\diamondsuit$ Georgia Institute of Technology\\
  \texttt{\{hui\_chen, wei\_han\}@mymail.sutd.edu.sg}\\
  \texttt{diyi.yang@cc.gatech.edu}\\
  \texttt{sporia@sutd.edu.sg}\\
  }

\begin{document}
\maketitle
\begin{abstract}
This paper proposes a simple yet effective interpolation-based data augmentation approach termed \textsc{DoubleMix}, to improve the robustness of models in text classification. \textsc{DoubleMix} first leverages a couple of simple augmentation operations to generate several perturbed samples for each training data, and then uses the perturbed data and original data to carry out a two-step interpolation in the hidden space of neural models. Concretely, it first mixes up the perturbed data to a synthetic sample and then mixes up the original data and the synthetic perturbed data. \textsc{DoubleMix} enhances models' robustness by learning the ``shifted'' features in hidden space. On six text classification benchmark datasets, our approach outperforms several popular text augmentation methods including token-level, sentence-level, and hidden-level data augmentation techniques. Also, experiments in low-resource settings show our approach consistently improves models' performance when the training data is scarce. Extensive ablation studies and case studies confirm that each component of our approach contributes to the final performance and show that our approach exhibits superior performance on challenging counterexamples. Additionally, visual analysis shows that text features generated by our approach are highly interpretable. Our code for this paper can be found at \url{https://github.com/declare-lab/DoubleMix.git}.
\end{abstract}

\section{Introduction}

Deep neural networks have enabled breakthroughs in most supervised settings in natural language processing (NLP) tasks. However, labeled data in NLP is often scarce, as linguistic annotation usually costs large amounts of time, money, and expertise. With limited training data, neural models will be vulnerable to overfitting and can only capture shallow heuristics that succeed in limited scenarios, which will lead to severe performance degradation when applied to challenging situations. 

In order to improve the robustness of models, various data augmentation methods have been proposed. Generally, there are three types of augmentation techniques: token-, sentence-, and hidden-level transformation. 
\citet{wei2019eda} summarized several common token-level transformations, including word insertion, deletion, replacement, and swap. Sentence-level transformation is to paraphrase a sentence through specific grammatical or syntactic rules. Back-translation~\citep{sennrich2016improving,edunov2018understanding} is a typical sentence-level augmentation method where a sentence is translated to an intermediate language and then translated back to obtain augmented samples. Additionally, for natural language inference (NLI) tasks that identify whether a premise entails, contradicts, or is neutral with a hypothesis, \citet{min2020syntactic} studied syntactic rules of sentences in inference tasks and proposed several syntactic transformation techniques such as \textit{Inversion} and \textit{Passivization} to construct syntactically informative examples. However, these methods often have high requirements for sentence structures. It is hard to obtain a large number of augmented samples by this method.

In recent years, several hidden-level augmentation methods are proposed and they have exhibited superior performance in a number of popular text classification tasks. TMix~\citep{chen2020mixtext} is a typical approach where a linear interpolation is performed in the hidden space of transformer models such as BERT~\citep{devlin2019bert}. The main idea of TMix~\citep{verma2019manifold} comes from Mixup, a method that is based on the principle of Vicinal Risk Minimization (VRM)~\citep{chapelle2001vicinal} and has achieved substantial improvements in computer vision tasks~\citep{verma2019manifold,hendrycks2020augmix,kim2020puzzle,rame2021mixmo} and natural language tasks~\citep{guo2019mixup,chen2020mixtext,kim2021linda,park2022data}. Recently, SSMix~\cite{yoon2021ssmix} which interpolates text based on the saliency of tokens~\cite{simonyan2014deep} in hidden space has been introduced. 
These methods make models learn a mapping from a mixed text representation to an intermediate label which is generated by linearly combining two different source labels. However, the intermediate soft label cannot always accurately describe the true probability of classes that the mixed text representations belong to, which limit the effectiveness of augmentation.

To overcome these limitations, this work proposes a simple yet effective interpolation-based data augmentation method termed \textsc{DoubleMix}, which performs interpolation in the hidden space and does not require label mixing. Firstly, we leverage a collection of simple augmentation operations to generate several perturbed samples from the raw data and then mix up these perturbed samples. Secondly, we mix up the original data with the synthesized perturbed data. We constrain the mixing weight of the original to be larger than the synthesized perturbed data to balance the trade-off between proper perturbations and the potential injected noise. To stabilize the training process, we add a Jensen-Shannon divergence regularization term to our training objective to minimize the distance between the predicted distributions of the original data and the perturbed variants.

To demonstrate the effectiveness of our approach, we conduct extensive experiments by comparing our \textsc{DoubleMix} with previous state-of-the-art data augmentation methods on six popular text classification benchmark datasets. 
Additionally, we reduce and vary the amount of training data, to observe if \textsc{DoubleMix} can consistently improve over the baselines. We further conduct ablation studies and case studies to investigate the impact of different training strategies on \textsc{DoubleMix}’s effectiveness and whether our method works on challenging counterexamples. Moreover, we visualize the features generated by \textsc{DoubleMix} to interpret why our method works. Experimental results and analyses confirm the efficacy of our proposed approach and every component in \textsc{DoubleMix} contributes to the performance. To sum up, our contributions are: 
\begin{itemize}
    \item We propose a simple interpolation-based data augmentation approach \textsc{DoubleMix} to improve the robustness of neural models in text classification by mixing up the original text with its perturbed variants in hidden space.
    \item We demonstrate the effectiveness of \textsc{DoubleMix} through extensive experiments and analyses on six text classification benchmarks as well as three low-resource datasets.
    \item We qualitatively analyze why our method works by visualizing its data manifold and quantitatively analyze how our method works by conducting several ablation studies and case studies.
\end{itemize}

\section{Related Work}

Data augmentation techniques are widely employed in NLP tasks to improve the robustness of models~\cite{sun2020mixup,xie2020unsupervised,cheng2020advaug,guo2020sequence,kwon2022explainability}. One way to enrich the original training set is to perturb the tokens in each sentence. For example, \citet{wei2019eda} introduced a set of simple data augmentation operations such as synonym replacement, random insertion, swap, and deletion. However, token-level perturbation sometimes does not guarantee that the augmented sentences are grammatically correct. 

Thus, sentence-level augmentation methods are introduced, where people paraphrase the sentence by some specific rules. \citet{minervini2018adversarially,mccoy-etal-2019-right} leveraged syntactic rules to generate adversarial examples in inference tasks. Moreover, \citet{andreas2020good} investigated the compositional inductive bias in sequence models and augmented data by compositional rules. 
However, these methods require careful design, and they are often customized for a specific task, which makes them hard to generalize to different datasets.

Recently, a couple of hidden-level augmentation techniques which perform interpolation in hidden space have been studied~\citep{guo2019mixup,verma2019manifold,hendrycks2020augmix}. Inspired by PuzzleMix~\cite{kim2020puzzle} and SaliencyMix~\cite{uddin2020saliencymix} which is popular in computer vision, \citet{yoon2021ssmix} proposed SSMix which utilizes the saliency information of spans~\cite{simonyan2014deep} in each sentence to interpolate in hidden space to create informative examples. \citet{yin2021batchmixup} interpolate hidden states of the entire mini-batch to obtain better representations. Inspired by the prior work, our \textsc{DoubleMix} aims at improving models' robustness by mixing up text features and their perturbed samples in hidden space.


\section{Proposed Method: \textsc{DoubleMix}}

To regularize NLP models in a more efficient manner, we introduce a simple yet effective data augmentation approach, \textsc{DoubleMix}, that enhances the representation of each training data by learning the features sampled from a region constructed by the original sample itself and its perturbed samples. The perturbed samples are generated by simple token- or sentence-level augmentation operations. \textsc{DoubleMix} is a hidden-level regularization technique and our base model is a pre-trained transformer network, as they have achieved great performance in various NLP tasks. \cref{alg:doublemix} shows the training process of our approach.

\subsection{Robust Interpolation in Hidden Space}
For an input sequence $x = \{w_0, w_1, ..., w_S\}$ with $S$ tokens associated with a label $y$, our goal is to predict a label of this sequence. At the beginning of this approach, we prepare a perturbation operation set containing simple token- and sentence-level data augmentation techniques such as back-translation (sentence-level), synonym replacement (token-level), and Gaussian noise perturbation (sentence-level). 
Thereafter, we randomly sample the operations $N$ times and use the selected augmentation operations to generate $N$ perturbed samples of each training instance. 
Note that each type of operation can be selected multiple times. 
We generated different perturbed samples by adjusting the hyper-parameters. 
For example, if we select synonym replacement, we can produce different perturbations by adjusting the proportion of tokens to be substituted. 
For back-translation, we can use different intermediate languages.

Our approach is performed in hidden space, so as to encourage the model to fully utilize the hidden information within the multi-layer networks. We employ a pre-trained model $f(;\theta)$ containing $L$ layers to encode the text to hidden representations. 
Then we select a layer $i$ which ranges in $[0, L]$ to interpolate. At the $i$-th layer of $f(;\theta)$, a two-step interpolation is performed where the first step is to mix up all the perturbed samples by a group of weights sampled from Dirichlet distribution, and the second step is to mix up the synthesized perturbed sample and the original sample by some weights $\in [0, 1]$ sampled from Beta distribution. We follow~\citet{zhang2018mixup} to use Beta distribution for weight sampling, and Dirichlet distribution is a multi-variate Beta distribution. Note that when we mix up the original data and the synthesized perturbed data, we constrain the mixing weight of the original data to be larger, so as to make the final perturbed representation to be close to the original one. This balances the trade-off between proper perturbation and potential injected noise. After the two-step interpolation, the synthesized hidden presentation is fed to the remaining layers $f_{[i:L]}(;\theta)$ and a classifier.

\begin{algorithm}[ht!]
\small
\caption{\textsc{DoubleMix}}\label{alg:doublemix}
\SetAlgoLined
\KwIn{Model $f(;\theta)$ containing $L$ layers, the $l$-th layer of the model $f_l(;\theta)$, classifier $\hat{p}(;\phi)$, training set $\mathcal{X}$ = \{$(x_1, y_1), ..., (x_n, y_n)$\}, perturbation operation set $O$ = \{back-translation (BT), synonym replacement (SR), ..., Gaussian noise (GN)\}, interpolation layer set $I = \{i_1, ..., i_k\}$, number of augmented samples $N$, number of training epochs $K$, the global learning rate $\eta$, Beta distribution hyper-parameter $\alpha$, Dirichlet distribution hyper-parameter $\tau$, loss hyper-parameter $\gamma$}
\BlankLine
\KwOut{Updated network weights $\theta$, $\phi$}
\BlankLine
$\{o_1, ..., o_N\}\gets O$ \Comment{Select $N$ operations. Each type of operation can be selected multiple times.}\\
\BlankLine
\For{$k\gets0$ \KwTo $K$}{
\BlankLine
    \For{$(x, y) \in \mathcal{X}$}{
    \BlankLine
    $\{x_a, ..., x_N\}\gets \{o_1, ..., o_N\}(x)$ \Comment{Apply the selected operations to generate $N$ different augmented samples of $x$.} \\
    \BlankLine
    $i \gets I = \{i_1, ..., i_k\} $ \Comment{Randomly select an interpolation layer from Set $I$}
    \BlankLine
    $h_{orig}^i \gets f_{[0:i]}(x;\theta)$ \\
    \BlankLine
    $\{h_{a}^i, ..., h_{N}^i\} \gets f_{[0:i]}(\{x_a, ..., x_N\};\theta)$ \Comment{Encode \{$x, x_a, ..., x_N$\}, and interpolate at the $i$-th layer.}\\
    \BlankLine
    Sample $(w_1,...,w_n)\sim \mathrm{Dirichlet}(\tau,...,\tau)$ \\
    \BlankLine
    $h_{aug}^i \gets w_1\cdot h_a^i+\ldots + w_n\cdot h_N^i$ \Comment{First mixup (\emph{Step \uppercase\expandafter{\romannumeral1}}).} \\
    \BlankLine
    Sample $\lambda \sim \mathrm{Beta}(\alpha, \alpha)$ \\
    \BlankLine
    $\lambda \gets \mathrm{max}{(\lambda, 1-\lambda)}$ \Comment{Constrain the synthetic data to a region closer to the original example.}\\
    \BlankLine
    $h_{mix}^i \gets \lambda \cdot h_{orig}^i + (1-\lambda) \cdot h_{aug}^i $ \Comment{Second mixup (\emph{Step \uppercase\expandafter{\romannumeral2}}).} \\
    \BlankLine
    $h_{mix} \gets f_{[i+1:L]}(h_{mix}^i;\theta)$
    \BlankLine
    $p_{mix} \gets \hat{p}(y|h_{mix};\phi)$\\
    \BlankLine
    $p_{orig} \gets \hat{p}(y|f(x;\theta);\phi)$\\
    \BlankLine
    $\bar{p} \gets \frac{1}{2} (p_{mix} + p_{orig})$ \\
    \BlankLine
    $\mathcal{L}_{JSD}(y|x, x_a, ..., x_N) \gets \frac{1}{2}(\mathrm{KL}(p_{mix}||\bar{p}) + \mathrm{KL}(p_{orig}||\bar{p}))$ \\
    \BlankLine
    $\mathcal{L} \gets \mathcal{L}_{CE}(y|x) + \gamma \mathcal{L}_{JSD}(y|x, x_a, ..., x_N)$ \\
    \BlankLine
    $\theta \gets \theta - \eta \nabla_{\theta, \phi}\mathcal{L}; \phi \gets \phi - \eta \nabla_{\theta, \phi}\mathcal{L}$\Comment{Update the network weight $\theta$ of the base model $f$ and $\phi$ of the classifier $\phi$.}
    }
}
\end{algorithm}

\subsection{Training Objectives}
During the training period, we do not directly minimize the Cross-Entropy loss of the probability distribution of the synthesized sample, as it may introduce too much noise. We employ a consistency regularization term, Jensen-Shannon Divergence (JSD) loss~\citep{Bachman2014pseudo,zheng2016improving,hendrycks2020augmix}, to minimize the difference between the prediction distribution of synthetic data and original data, and meanwhile, we minimize the Cross-Entropy loss of the model output of original data and the gold label. The training objective can be written as:
\begin{flalign}
\mathcal{L}(y|x,x_{a},...,x_{N}) = \mathcal{L}_{CE}(y|x) \nonumber\\
+ \gamma \mathcal{L}_{JSD}(y|x,x_{a},...,x_{N}) 
\end{flalign}
where $\gamma$ is a hyper-parameter. 

In the consistency regularization, we do not employ Kullback-Leibler divergence (KL) because it is not symmetric, i.e., $\KL(P||Q) \neq \KL(Q||P)$ when $P \neq Q$. It is not a promising choice to measure the similarity of $p_{mix}$ and $p_{orig}$ using KL, as neither $p_{mix}$ nor $p_{orig}$ are true predictions and we deem that they share equal status. JSD provides a smoothed and normalized version of KL divergence, with scores between 0 (identical) and 1 (maximally different). We believe using such a symmetric metric can make the training more stable. 

\subsection{Why does \textsc{DoubleMix} work?}
\begin{figure}[ht]
    \centering
    \includegraphics[width=0.96\columnwidth]{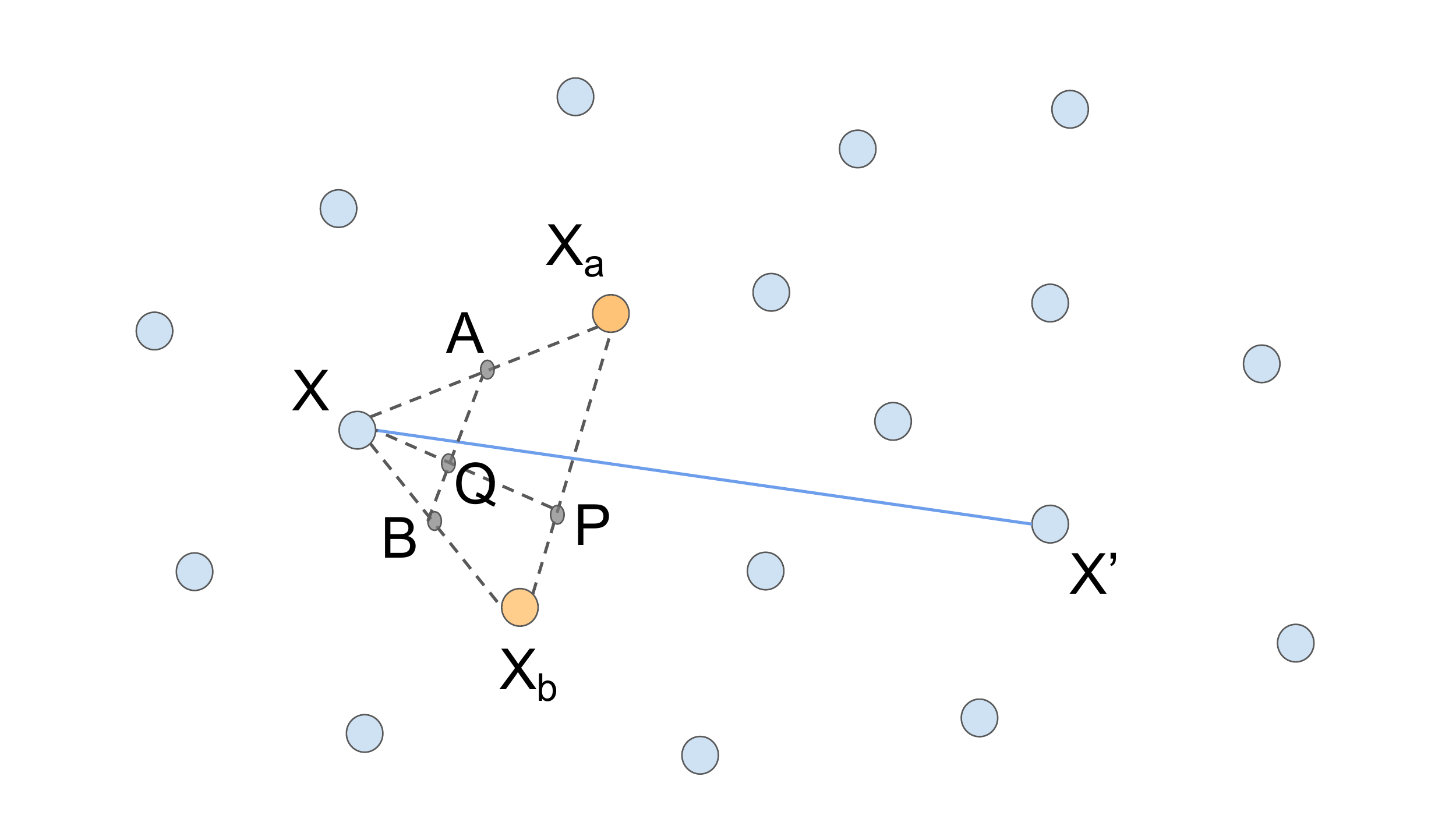}
    \caption{Visualization of the sample space of synthesized data in Mixup and \textsc{DoubleMix}. Blue dots $X$ and $X'$ indicate two data points in the training set. Orange dots $X_a$ and $X_b$ are perturbed data of $X$. Grey dots $A$, $B$, $P$ and $Q$ are sampled points.
    }
    \label{fig:manifold}
\end{figure}

To further discuss why our method works, we visualize the original data and the sample space of synthesized data in Mixup and \textsc{DoubleMix} in~\cref{fig:manifold}. For brevity, we assume the number of perturbed samples in Step \uppercase\expandafter{\romannumeral1} is two. As shown in~\cref{fig:manifold}, blue dots indicate training data, and orange dots are perturbed data generated by our selected operations. Synthesized data in Mixup~\citep{zhang2018mixup} can only be created along a line, such as the blue full line connecting the two points $X$ and $X'$ since it is a simple linear combination. Hence, Mixup enforces the regularization to behave linearly among the training data. In contrast to Mixup, the sample space of synthesized data of \textsc{DoubleMix} is a polygon. In this example, $\triangle XAB$ is the sample space of the synthesized data in \textsc{DoubleMix}. Firstly, Step \uppercase\expandafter{\romannumeral1} samples a point $P$ on the line connecting $X_a$ and $X_b$. Secondly, Step \uppercase\expandafter{\romannumeral2} finds a point $Q$ on the line connecting $X$ and $P$. Note that $Q$ should be closer to $X$ or at the middle of Line $XP$, as in Step \uppercase\expandafter{\romannumeral2}, we constrain the mixing weight of original data to be larger than that of synthesized perturbed data. Taken together, our approach enforces the model to learn nearby features for each training data so that it is robust to representation shifts.


\section{Experimental Setup}


\begin{table*}[ht]
    \centering
    \scalebox{0.86}{
    \begin{tabular}{l|cc|cc|cc}
    \toprule
      \multirow{2}{*}{\textbf{Method}} & \multicolumn{2}{c|}{\textbf{SST-2}} & \multicolumn{2}{c|}{\textbf{TREC}} & \multicolumn{2}{c}{\textbf{Puns}} \\ 
      & \textbf{Acc.} & \textbf{F1.} & \textbf{Acc.} & \textbf{F1.} & \textbf{Acc.} & \textbf{F1.} \\
      \midrule
    BERT~\citep{devlin2019bert} & $91.08_{0.1}$ & $91.09_{0.1}$ & $96.90_{0.4}$ & $96.21_{0.7}$ & $94.20_{0.2}$ & $94.38_{0.2}$\\
    + Easy Data Augmentation~\citep{wei2019eda} & $91.66_{0.2}$&$91.75_{0.1}$ & $97.10_{0.3}$& $96.74_{0.2}$& $93.71_{0.6}$& $93.83_{0.5}$\\
    + Back Translation~\citep{edunov2018understanding} & $91.33_{0.2}$& $91.29_{0.1}$& $96.90_{0.4}$&$96.72_{0.1}$ & $93.27_{0.4}$ & $93.34_{0.4}$\\
    + Manifold Mixup~\citep{verma2019manifold} & $91.33_{0.4}$& $91.44_{0.3}$& $97.00_{0.3}$& $96.29_{0.5}$& $94.37_{0.9}$& $94.44_{1.0}$\\
    + TMix~\citep{chen2020mixtext} & $91.13_{0.5}$& $91.29_{0.3}$&$96.90_{0.1}$ &$96.32_{0.3}$ & $94.21_{0.7}$&$94.28_{0.8}$ \\
    + SSMix~\citep{yoon2021ssmix} & $91.45_{0.5}$& $91.50_{0.2}$& $97.00_{0.2}$& $96.44_{0.2}$& $94.33_{0.7}$&$94.46_{0.6}$ \\
   
    + \textsc{DoubleMix} (Ours) &$\textbf{92.21}_{0.1}$ &$\textbf{92.11}_{0.1}$ & $\textbf{97.40}_{0.1}$& $\textbf{97.32}_{0.6}$&$\textbf{94.59}_{0.1}$ &$\textbf{94.66}_{0.1}$ \\
    \bottomrule
    \toprule
    \multirow{2}{*}{\textbf{Method}} & \multicolumn{2}{c|}{\textbf{IMDB}} & \multicolumn{2}{c|}{\textbf{SNLI}} & \multicolumn{2}{c}{\textbf{MNLI}} \\ 
      & \textbf{Acc.} & \textbf{F1.} & \textbf{Acc.} & \textbf{F1.} & \textbf{Acc.} & \textbf{F1.} \\
      \midrule
    BERT~\citep{devlin2019bert} & $83.52_{0.1}$ & $83.65_{0.4}$ & $90.31_{0.5}$ & $90.28_{0.5}$ & $84.16_{0.2}$ & $84.05_{0.2}$\\
    + Easy Data Augmentation~\citep{wei2019eda} & $83.43_{0.1}$& $83.85_{0.1}$& $90.27_{0.4}$&$90.28_{0.4}$ & $84.04_{0.1}$&$83.97_{0.2}$ \\
    + Back Translation~\citep{edunov2018understanding} &$83.69_{0.1}$ &$84.11_{0.2}$ &$90.21_{0.5}$ &$90.23_{0.4}$ & $84.50_{0.3}$ & $84.44_{0.3}$ \\
    + Syntactic Data Augmentation~\citep{min2020syntactic} & - & - &$90.35_{0.3}$ & $90.31_{0.4}$ & $84.19_{0.3}$ & $84.08_{0.2}$ \\
    + Manifold Mixup~\citep{zhang2018mixup} & $83.63_{0.1}$&$83.81_{0.2}$ & $90.04_{0.2}$ & $90.02_{0.2}$& $83.37_{0.2}$ & $83.31_{0.1}$ \\
    + TMix~\citep{chen2020mixtext} & $83.47_{0.3}$&$83.91_{0.2}$ & $90.12_{0.1}$&$90.09_{0.1}$ &$83.43_{0.1}$ &$83.38_{0.1}$ \\
    + SSMix~\citep{yoon2021ssmix} & $83.55_{0.3}$&$83.88_{0.2}$ & $90.21_{0.2}$& $90.14_{0.3}$ & $83.66_{0.2}$& $83.54_{0.2}$ \\
    + \textsc{DoubleMix} (Ours) & $\textbf{84.14}_{0.5}$ & $\textbf{84.39}_{0.2}$& $\textbf{91.03}_{0.1}$& $\textbf{91.02}_{0.1}$ & $\textbf{84.72}_{0.2}$ & $\textbf{84.64}_{0.1}$ \\
    \bottomrule
    \end{tabular}
    }
    \caption{Test accuracy (\%) and F1 scores (\%) for BERT when comparing our proposed \textsc{DoubleMix} with baseline methods on six text classification datasets. We randomly select two augmented samples to mix up in Step \uppercase\expandafter{\romannumeral1}.  Best scores are marked in bold. 
    As syntactic transformations~\citep{min2020syntactic} are rule-based data augmentation techniques customized for inference tasks, we only show their performances on SNLI and MNLI. We report the mean accuracy and F1 scores across five different runs with the standard deviation shown in subscript (e.g., $91.08_{0.1}$ indicates $91.08\pm0.1$). 
    }
    \label{tab:main_results}
\end{table*}

\subsection{Datasets}
We compare our approach with several data augmentation baselines on six text classification datasets, covering sentiment polarity classification, question type classification, humor detection, and natural language inference: IMDB~\citep{maas2011learning} and SST-2~\citep{socher2013recursive} which predict the sentiment of movie reviews to be positive or negative, 6-class open-domain question classification TREC~\citep{li2002learning}, Pun of the day (Puns)~\citep{yang2015humor} which detects humor in a single sentence, and two inference datasets SNLI~\citep{bowman2015large} and MNLI~\citep{williams2018broad} which identify whether a premise entails, contradicts or is neutral with a hypothesis. Statistics of the six text classification datasets can be found in~\cref{app:dataset}. As the test set of MNLI is not publicly available, we used the matched development set as our development set and the mismatched development set as our test set in our experiments.

\subsection{Baselines}
We compare our approach with several widely-used baseline models, including token-, sentence-, and hidden-level augmentation techniques. Token-level baselines contain the operations in Easy Data Augmentation (EDA)~\citep{wei2019eda} where they randomly insert, swap, and delete tokens in each sentence. Sentence-level baselines include paraphrasing sentences such as back-translation~\citep{sennrich2016improving} and applying some syntactic rules to create augmented sentences such as syntactic transformation~\citep{min2020syntactic}. Hidden-level baselines include Manifold Mixup~\citep{verma2019manifold}, TMix~\citep{chen2020mixtext}, and SSMix~\cite{yoon2021ssmix}, where they mix up two different training samples in hidden space and learn a mapping from the intermediate representation to an intermediate label. Details of the implementation of the baselines can be found in~\cref{app:baseline}.

\begin{table*}[ht]
    \centering
    \scalebox{0.65}{
    \begin{tabular}{l|cccc|cccc|c}
    \toprule
     \multirow{3}{*}{\textbf{Method}} & \multicolumn{4}{c|}{\textbf{SNLI}} & \multicolumn{4}{c|}{\textbf{MNLI}} & \\
      & \textbf{1K} & \textbf{2.5K} & \textbf{5K} & \textbf{10K}  & \textbf{1K} & \textbf{2.5K} & \textbf{5K} & \textbf{10K} & \textbf{Avg.}\\
      & (Acc./F1.) & (Acc./F1.)& (Acc./F1.) & (Acc./F1.)  & (Acc./F1.) & (Acc./F1.) & (Acc./F1.) & (Acc./F1.) & (Acc./F1.)\\
      \midrule
       BERT  & 69.77/69.59 & 76.10/75.92 & 79.28/79.25 & 82.36/82.28 & 55.81/54.61 & 65.63/65.15 & 71.24/71.01 & 74.24/74.14 & 71.80/71.49\\
       + BT &70.23/69.96 &76.51/76.54 & 79.57/79.57& 82.68/82.65& 57.28/55.53 & 66.97/66.95& 72.28/72.01& 74.49/74.44& 72.50/72.21\\
       + M-Mix & 71.45/71.34 & 76.48/76.42 & 79.91/79.83 & 82.14/82.13 & 57.01/56.70 & 67.08/66.96 & 71.76/71.68 & 74.68/74.54 & 72.56/72.45\\
       + TMix & 71.04/71.12 & 76.38/76.10 & 79.85/79.81 & 82.09/82.09 & \colorbox{yellow}{\textbf{57.31}}/56.99 & 67.10/67.01 & 71.66/71.59 & 74.86/74.70 & 72.54/72.43\\
       + SSMix  & 71.32/71.21& 76.87/76.72& 80.02/79.93& 82.41/82.20& 57.25/\colorbox{yellow}{\textbf{57.10}}& 67.13/67.05& 71.70/71.63& 74.77/74.62& 72.68/72.56\\
       + Ours & \colorbox{yellow}{\textbf{71.82}/\textbf{71.72}} & \colorbox{yellow}{\textbf{77.43}/\textbf{77.42}} & \colorbox{yellow}{\textbf{80.75}/\textbf{80.72}} & \colorbox{yellow}{\textbf{83.18}/\textbf{83.26}} &  56.15/55.91&  \colorbox{yellow}{\textbf{67.57}/\textbf{67.33}}& \colorbox{yellow}{\textbf{72.35}/\textbf{72.15}} & \colorbox{yellow}{\textbf{75.07}/\textbf{74.97}} & \colorbox{yellow}{\textbf{73.04}/\textbf{72.94}}\\
       \midrule
       RoBERTa & 78.11/77.95 & 82.47/82.30 & 83.17/83.36 & 85.58/85.50 & 70.23/70.45 & 75.50/75.53 & 79.01/79.04 & 81.00/80.98& 79.38/79.39 \\
       + BT & 78.32/78.25& 82.47/82.61& 84.23/84.08& 86.07/86.05& 70.67/70.52& \colorbox{yellow}{\textbf{77.58}/\textbf{77.39}}& 79.17/79.07& 81.26/81.08& 79.97/79.88\\
       + M-Mix & 79.32/78.73 & 82.71/82.63 & 84.60/84.63 & 86.00/85.95 & 71.78/71.04 & 76.00/75.96 & 79.43/79.34 & 81.40/81.28 &80.16/79.95\\
       + TMix & 79.17/79.11 & 82.84/82.91 & 85.09/85.13 & 86.16/86.13 & \colorbox{yellow}{\textbf{72.05}/\textbf{72.18}} & 76.57/76.44 & 79.92/79.80 & 81.30/81.17 &80.39/80.36 \\
       + SSMix  & 79.43/79.35& 82.91/82.88& 85.33/85.36& 86.32/86.28& 71.96/71.88& 76.44/76.38& 79.86/79.77& 81.25/81.22& 80.44/80.39\\
       + Ours & \colorbox{yellow}{\textbf{80.41}/\textbf{80.31}} & \colorbox{yellow}{\textbf{83.96}/\textbf{83.92}} & \colorbox{yellow}{\textbf{85.42}/\textbf{85.42}} & \colorbox{yellow}{\textbf{86.91}/\textbf{86.88}}& 71.20/71.15  & 77.12/76.96 & \colorbox{yellow}{\textbf{80.43}/\textbf{80.28}} & \colorbox{yellow}{\textbf{82.46}/\textbf{82.24}} & \colorbox{yellow}{\textbf{80.99}/\textbf{80.90}} \\
    \bottomrule
    \end{tabular}
    }
    \caption{Test accuracy (\%) and F1 score (\%) comparison on the SNLI and MNLI datasets training with varying amounts of training data (1000, 2500, 5000, and 10000). Best scores are marked in bold in yellow background. BT and M-Mix represent Back Translation and Manifold Mixup. We only use BT operations in \textsc{DoubleMix} in this experiment. 
    }
    \label{tab:low-data-results}
\end{table*}

\section{Results and Analysis}
We evaluate our baselines and proposed approach on six text classification benchmark datasets. 
We also show the performance of our approach in low-resource settings to confirm \textsc{DoubleMix} is efficient and robust when the training samples are scarce. This section will discuss the performance of models in detail and quantitatively analyze how and why \textsc{DoubleMix} works.

\subsection{Main Results}
\cref{tab:main_results} shows the performance of \textsc{DoubleMix} and the relevant baselines on six text classification datasets. Our base model is the BERT-base-uncased model. We observe that \textsc{DoubleMix} achieves the best average results compared to previous state-of-the-art baselines across six datasets, where \textsc{DoubleMix} shows the greatest improvements over BERT on SST-2, increasing the test accuracy and binary F1 score by 1.13\% and 1.02\%.

In addition, we find token-level Easy Data Augmentation and sentence-level Back Translation are not able to improve the BERT baseline on Puns and SNLI. Especially on Puns, Back Translation's test accuracy and F1 score are about 1\% lower than BERT. This might be that labels in humor detection and inference tasks are closely related to the presence of some important words, and Easy Data Augmentation and Back Translation may perturb these words, making the true label flip, but the label learned by the model does not change accordingly, which leads to an inefficient learning process. Moreover, hidden-level augmentations such as Manifold Mixup, TMix, and SSMix fail to improve the base model on SNLI and MNLI. As subtle changes in the sentences in inference tasks will flip the true label, mixing up two different samples and learning an intermediate representation in hidden space cannot ensure the learned soft label is the true label of the intermediate representation. 

In contrast, our model shows consistent improvements over BERT on these datasets. 
The consistent improvements indicate that, by strategically mixing up samples with similar meanings in the hidden space, \textsc{DoubleMix} not only helps pre-trained models to become insensitive to feature perturbations in an effective way but also injects less potential noise during the augmentation process compared to other baselines. 

\subsection{Performance in Low-Resource Settings}
To investigate the performance of our approach in low-resource settings, we randomly sample 1000, 2500, 5000, and 10000 examples from the original training data of SNLI and MNLI to construct our training sets for low-data setting evaluations, while the size of the development and test sets is unchanged. 
Apart from the BERT-base-uncased model~\citep{devlin2019bert}, we also conduct experiments on the RoBERTa-base model~\citep{liu2019roberta}. 

\cref{tab:low-data-results} presents the results in low-data settings. Compared with the BERT baseline in~\cref{tab:main_results}, we can observe that although pre-trained language models are powerful across text classification tasks, the test accuracy and F1 scores might decrease a lot when the training data is very scarce. \textsc{DoubleMix} consistently improves the base model with no data augmentation on both SNLI and MNLI and outperforms all the baselines on the SNLI dataset. 
On MNLI, we observe that our method always achieves the top performance except in the 1K training samples.
As the training set grows larger, our model gradually outperforms the baselines and the leading gap keeps expanding---when the number of training samples reaches 10K, our model can achieve at least 1\% higher accuracy and F1 score than RoBERTa on both SNLI and MNLI.

\subsection{Ablation Studies}
\subsubsection{Training Strategies in \textsc{DoubleMix}}
We also conduct a series of ablative experiments to examine the contribution of individual components.
The results are displayed in~\cref{tab:ablation_study}, where our experiments are conducted on the SNLI dataset with only 1000 training samples. We find the performance drops after changing the training strategies, suggesting that the current interpolation method trained with JSD loss in \textsc{DoubleMix} contributes to the final performance. 

Concretely, we first remove the JSD loss in our training objective to check if this loss contributes to the performance. 
We observe that the accuracy and F1 score drop approximately 0.7\% and 0.9\% after removing the JSD loss, which manifests that JSD loss is capable of stabilizing the training process. 
Secondly, we merge the two steps and see how the model performs, where we use a Dirichlet distribution to sample $N+1$ mixing weights for the original example and other $N$ augmented examples, and mix up them at a time. In this case, the test accuracy and F1 score drop to 71.11\% and 70.96\%, respectively. 
This indicates that the two-step interpolation where we constrain the synthetic data to a region closer to the original sample, as mentioned in Line $11$ in~\cref{alg:doublemix} will inject less noise. 
Moreover, we have also tried different mixing samples Step \uppercase\expandafter{\romannumeral2}. 
We find mixing up with another randomly selected training sample in Line $12$ in~\cref{alg:doublemix} results in a 0.59\% accuracy decrease and a 0.70\% F1 score decrease. 
If the selected training sample is restricted to the same category as the original data, the performance degradation will be even larger. 
This outcome may be caused by the larger semantic difference between the selected example and the original data compared to the augmented examples of the original data.
\begin{table}[htbp]
    \centering
    \scalebox{0.72}{
    \begin{tabular}{l|cc}
    \toprule
    \textbf{Method} & \textbf{Acc.} & \textbf{F1.}\\
    \midrule
        \textsc{DoubleMix} & 71.82 & 71.72\\
        - w/o JSD loss & 71.15 & 70.87\\
        - merge Step \uppercase\expandafter{\romannumeral2} and Step \uppercase\expandafter{\romannumeral1}  & 71.11 & 70.96\\
        - mix with another training sample in Step \uppercase\expandafter{\romannumeral2}& 71.23 & 71.02\\
        - mix with another same-class sample in Step \uppercase\expandafter{\romannumeral2}& 70.98 & 70.74\\
    \bottomrule
    \end{tabular}}
    \caption{Test accuracy (\%) and F1 scores (\%) on the SNLI dataset with 1000 training samples after changing different parts of \textsc{DoubleMix}.}
    \label{tab:ablation_study}
\end{table}

\begin{table}[h]
    \centering
    \scalebox{0.9}{
    \begin{tabular}{c|cccc}
    \toprule
      \textbf{Mixup layer set}  & \textbf{Acc.} &$\Delta_{Acc}$& \textbf{F1.} &$\Delta_{F1}$\\
    \midrule
      $\emptyset$ & 69.77& &69.59& \\
      \{0\} & 71.13& +1.36 &71.02&+1.43\\
      \{0,1,2\} & 70.95&+1.18&70.84&+1.25\\
      \{3,4\} & 71.56 &+1.79 &71.48 & +1.89 \\ 
      \{3,6,9\} & 71.24&+1.47& 71.16&+1.57\\
      \{7,9,12\} & 71.30&+1.53& 71.19& +1.60 \\
      \{9,10,12\} & \textbf{71.82} &\textbf{+2.05} & \textbf{71.72} & \textbf{+2.13}\\
      \{3,4,6,9,10,12\} & 70.90 &+1.13 & 70.90 & +1.31 \\
      \{3,4,6,7,9,12\} & 71.29 &+1.52 & 71.16 & +1.57 \\
    \bottomrule
    \end{tabular}
    }
    \caption{Test accuracy (\%) and F1 scores (\%) on SNLI with 1000 training data with different interpolation layer sets. $\emptyset$ means no interpolation, and \{0\} is the input space. $\Delta$ indicates the gap to the baseline with no augmentation. }
    \label{tab:mix_layer_set_result}
\end{table}

\subsubsection{Effect of Interpolation Layers}
We believe the hidden layers in pre-trained language models are powerful in representation learning, and interpolation in the hidden space can yield a larger performance improvement than in the input space. 
In this section, we will investigate which interpolation option in terms of the set of layers in pretrained models can obtain the best performance. 
Previous work \citet{jawahar2019bert} indicates that BERT's intermediate layers \{3, 4\} perform best in encoding surface features and layers \{6, 7, 9, 12\} contain the most syntactic features and semantic features. We refer to~\citet{jawahar2019bert} to formulate several sets of layers and have conducted a couple of additional experiments on BERT + \textsc{DoubleMix} with different sets of interpolation layers on SNLI with 1000 training samples to see which subsets give the optimal performance. The results are shown in~\cref{tab:mix_layer_set_result}. 

\begin{table}[ht]
    \centering
    \scalebox{0.73}{
    \begin{tabular}{l|l|c|c}
    \toprule
     \textbf{Model} & \textbf{Original and Counterfactual Examples}&\textbf{P.}&\textbf{T.}\\
    \midrule
       \multirow{2}{*}{BERT} & \textit{P}: Students are inside of a lecture hall. & \multirow{2}{*}{\color{red}N} & \multirow{2}{*}{E}\\
         & \textit{H}: Students are indoors. & &   \\
    \midrule
        \multirow{2}{*}{Ours} & \textit{P}: Students are inside of a lecture hall. & \multirow{2}{*}{\color{blue}E} & \multirow{2}{*}{E}\\
        & \textit{H}: Students are indoors. & &  \\
    \midrule
    \midrule
       \multirow{2}{*}{BERT} & \colorbox{yellow}{\textit{P}: Students are inside of a lecture hall.} & \multirow{2}{*}{\color{blue}C} & \multirow{2}{*}{C}\\
         & \colorbox{yellow}{\textit{H}: Students are on the soccer field.} & &   \\
    \midrule
        \multirow{2}{*}{Ours} & \colorbox{yellow}{\textit{P}: Students are inside of a lecture hall.} & \multirow{2}{*}{\color{blue}C} & \multirow{2}{*}{C}\\
        & \colorbox{yellow}{\textit{H}: Students are on the soccer field.} & &  \\
    \midrule
     \midrule
       \multirow{2}{*}{BERT} & \textit{P}: Man in green jacket with baseball hat on. & \multirow{2}{*}{\color{blue}C} & \multirow{2}{*}{C}\\
         & \textit{H}: The man is not wearing a hat. & &   \\
    \midrule
        \multirow{2}{*}{Ours} & \textit{P}: Man in green jacket with baseball hat on. & \multirow{2}{*}{\color{blue}C} & \multirow{2}{*}{C}\\
        & \textit{H}: The man is not wearing a hat. & &  \\
    \midrule
    \midrule
       \multirow{2}{*}{BERT} & \colorbox{yellow}{\textit{P}: Man in green jacket with baseball hat on.} & \multirow{2}{*}{\color{red}E} & \multirow{2}{*}{N}\\
         & \colorbox{yellow}{\textit{H}: The man is at a baseball game.} & &   \\
    \midrule
        \multirow{2}{*}{Ours} & \colorbox{yellow}{\textit{P}: Man in green jacket with baseball hat on.} & \multirow{2}{*}{\color{blue}N} & \multirow{2}{*}{N}\\
        & \colorbox{yellow}{\textit{H}: The man is at a baseball game.} & &  \\
    \bottomrule
    \end{tabular}}
    \caption{Predictions of  BERT and our method on original examples and their counterfactual examples on the SNLI dataset. The counterfactual examples are extracted from \citet{Kaushik2020Learning} and are constructed by substituting entities or adding details to entities. The examples in the yellow background are counterexamples. \textbf{P.} and \textbf{T.} represent prediction and true label. E, N and C are \textit{entailment}, \textit{neutral}, and \textit{contradiction}. \textit{P} and \textit{H} are premise and hypothesis. Labels in red are wrong predictions while labels in blue are correct predictions.}
    \label{tab:case_study}
\end{table}
\begin{figure*}
    \centering
    \scalebox{0.9}{
    \includegraphics[width=\textwidth]{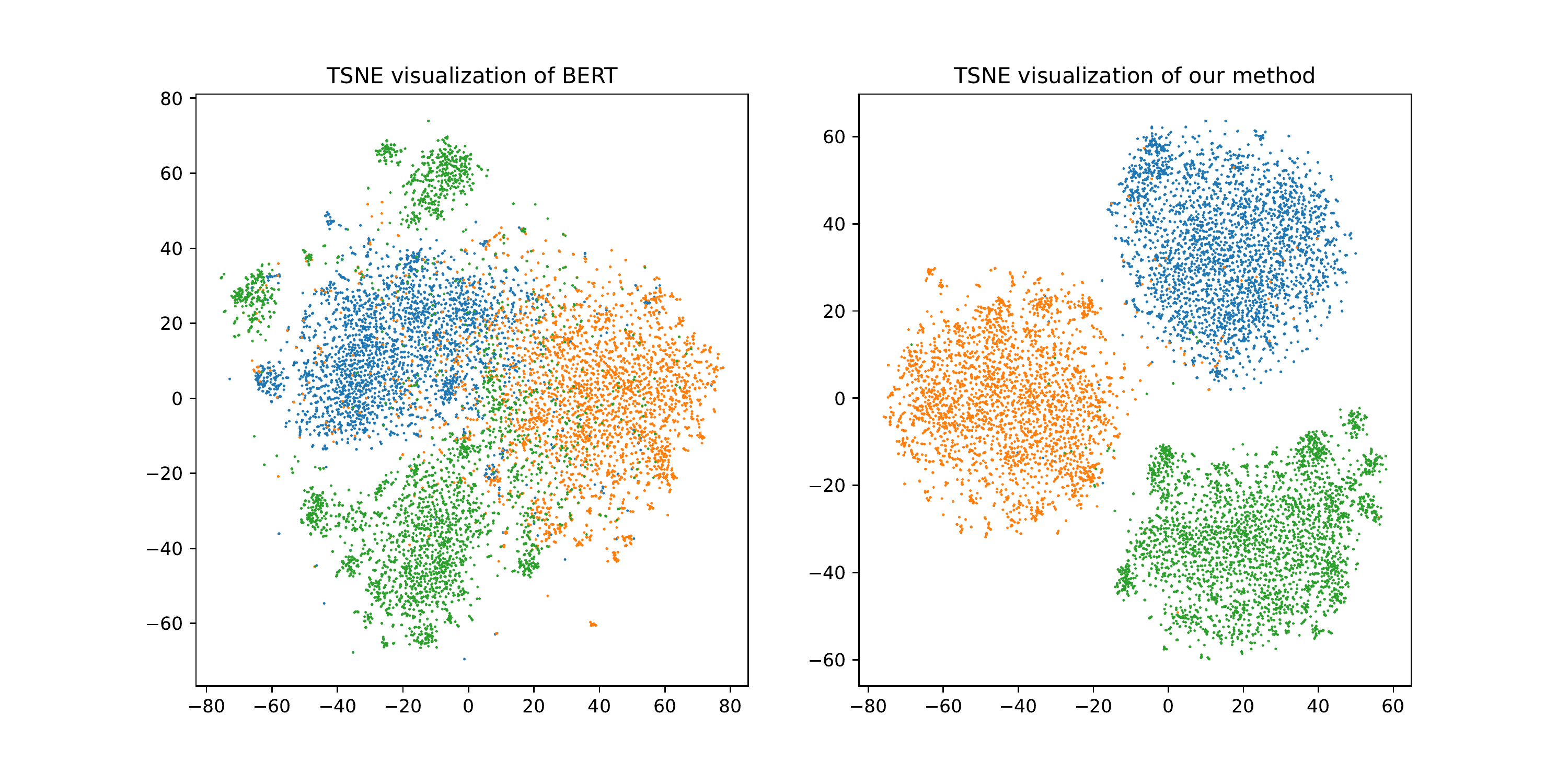}}
    \caption{T-SNE projection of the features generated by the 12-layer encoder of BERT baseline (left) and BERT + \textsc{DoubleMix} (right) on SNLI with 10000 training samples. The visualized features of the augmented text are extracted from the last layer of the base model during testing.}
    \label{fig:manifold_vis}
\end{figure*}

When all the interpolation steps are excluded, the test accuracy is 69.77\% and the F1 score is 69.59\%.
When we interpolate in the input space (the 0-th layer), the accuracy and F1 score increase by about 1.3\% and 1.4\%, showing that interpolation contributes to the performance. 
When we perform interpolations at layer set \{0, 1, 2\} with lower layers, the accuracy and F1 increases are smaller than interpolating in input space. 
However, when we interpolate at some middle layers such as \{3, 4\} and \{3, 6, 9\}, the performance improvement is more significant. 
According to~\citet{jawahar2019bert}, the 9-th layer captures most of the syntactic and semantic information. 
We have tried several layer sets containing the 9-th layer and find \{9, 10, 12\} containing upper layers performs best. 
At the same time, we notice that the number of layers in the interpolation layer set is not the more the better. The performance of \{3,4,6,9,10,12\} is only 70.90\%, which is lower than any other layer set, and the performance of \{3,4,6,7,9,12\} is not the best, indicating that too many interpolation layers will reduce the efficiency of representation learning.

\subsection{Case Studies}
To further understand how \textsc{DoubleMix} works, we randomly pick up some examples from the SNLI dataset and check the discrepancy between the predictions obtained from BERT and our method. 
Predictions of BERT and \textsc{DoubleMix} are shown in~\cref{tab:case_study}. We find in those examples with ``\textit{contradiction}'' label, both BERT and \textsc{DoubleMix} can accurately predict the true label. 
Additionally, to investigate how our method behaves in challenging scenarios, we also test on the counterfactual version~\cite{Kaushik2020Learning} of our selected samples.  
Both models excel in detecting negative words ``\textit{not}'' and location names. 
However, when the ground truth is ``\textit{entailment}'' or ``\textit{neutral}'', \textsc{DoubleMix} is more likely to make correct predictions. 
When the premise and hypothesis contain some common words (e.g., \textit{Man in green jacket with \textbf{baseball} hat on. The man is at a \textbf{baseball} game.}), \textsc{DoubleMix} inclines to make more accurate predictions.

\subsection{Manifold Visualization}
Finally, we visualize the embedding vectors generated by the 12-layer encoder of the BERT baseline with no data augmentation (BERT) and with \textsc{DoubleMix} to qualitatively show the effectiveness of our approach in facilitating the model to learn robust representations in~\cref{fig:manifold_vis}. Our experiments are conducted on SNLI with 10000 training samples. The visualized features are extracted from the output of the last layer of the model. We employ t-SNE~\cite{van2008visualizing} which is implemented by the python package scikit-learn\footnote{\url{https://github.com/scikit-learn/scikit-learn}} to visualize the features. In~\cref{fig:manifold_vis}, there are three clusters with three colors indicating different classes. We observe that the features in \textsc{DoubleMix} are better separated than those in BERT, indicating that our method effectively improves robustness by encouraging the model to learn nearby features of each training sample. 

\section{Conclusion}
In this work, we present a simple interpolation-based data augmentation approach \textsc{DoubleMix} to improve models' robustness on a wide range of text classification datasets. \textsc{DoubleMix} first leverages simple augmentation operations to generate perturbed data of each training sample and then performs a two-step interpolation in the hidden space of models to learn robust representations. Our approach outperforms several popular data augmentation methods on six benchmark datasets and three low-resource datasets. Finally, ablation studies, case studies, and visualization of manifold further explain how and why our method works. Our future work includes making the mixing weights learnable as well as extending \textsc{DoubleMix} to natural language generation tasks.

\section*{Acknowledgement}
This work is supported by the A*STAR under its RIE 2020 AME programmatic grant
RGAST2003 and project T2MOE2008 awarded by Singapore's MoE under its Tier-2 grant scheme.

\bibliography{main}
\bibliographystyle{acl_natbib}

\appendix
\section{Dataset Statistics}
\label{app:dataset}
\cref{tab:full-dataset} describes the statistics of the datasets we used. Note that for SST-2, we did not use the one on the GLUE benchmark, as the test labels are not publicly available. We used the original SST-2 dataset and it can be loaded from huggingface datasets\footnote{\url{https://huggingface.co/datasets/gpt3mix/sst2}}.
\begin{table}[ht]
    \centering
    \scalebox{0.83}{
    \begin{tabular}{l|ccc}
    \toprule
      \textbf{Dataset} & \textbf{Task Type} & \textbf{\# Label} & \textbf{Size} \\
      \midrule
       SST-2 & Sentiment & 2 & 6.9k / 872 / 1.8k\\
       TREC & Classification & 6 & 4.9k / 546 / 500\\
       Puns & Humor & 2 & 3.6k / 603 / 604\\
       IMDB & Sentiment & 2 & 22.5k / 2.5k / 25k\\
       SNLI & Inference & 3 & 550k / 9.8k / 9.8k\\
       MNLI & Inference & 3 & 392k / 9.8k / 9.8k\\
    \bottomrule
    \end{tabular}}
    \caption{Summary statistics of the seven natural language understanding datasets. We report the size of datasets as (train / validation / test) format.}
    \label{tab:full-dataset}
\end{table}

\section{Baseline Details}
\label{app:baseline}
This section introduces the details of our baselines:
\begin{itemize}
\item \textbf{Easy Data Augmentation}~\citep{wei2019eda} contains several simple data augmentation techniques in text such as synonym replacement, random insertion, random swap, and random deletion. The experimental setup for these methods is the same as that of back-translation. We use the official code\footnote{\url{https://github.com/jasonwei20/eda_nlp}} with the default insertion/deletion/swap ratio the author provided.

\item \textbf{Back-translation}~\citep{sennrich2016improving}
translates an input text in some source language (e.g. English) to another intermediate language (e.g. German), and then translates it back into the original one. In our experiments, our intermediate languages are German and Russian. And we create two types of augmented text for every training sample. These augmented examples are directly added to the training set. We use the code of fairseq\footnote{\url{https://github.com/pytorch/fairseq/blob/main/examples/wmt19/README.md}} to implement this baseline.

\item \textbf{Syntactic Transformation}~\citep{min2020syntactic}
applies rule-based syntactic transformations such as inversion and passivization to sentences to generate augmentations in inference tasks. We directly add the augmented data to the training set. The implementation is based on the official code\footnote{\url{https://github.com/aatlantise/syntactic-augmentation-nli}} the author provided.

\item \textbf{Manifold Mixup}~\citep{verma2019manifold} performs in hidden space. Similar to Mixup~\cite{zhang2018mixup}, Manifold Mixup samples two training examples and mixes up the hidden representations using a coefficient $\lambda_0$ randomly sampled from Beta($\alpha$, $\beta$). For the training objective, Manifold Mixup first uses the Cross-Entropy loss to measure the divergence between the predicted distribution and the one-hot vector of gold label, and then mix up the Cross-Entropy losses. Our implementation is based on the official code of Mixup\footnote{\url{https://github.com/hongyi-zhang/mixup}}.

\item \textbf{TMix}~\citep{chen2020mixtext} is similar to Manifold Mixup which performs interpolation in hidden space. We first mix up the gold labels to a sythetic label and secondly minimize the KL divergence between the synthetic label and the predicted distribution. We use the code implemented in the MixText\footnote{\url{https://github.com/GT- SALT/MixText}} repository.

\item \textbf{SSMix}~\cite{yoon2021ssmix} is similar to PuzzleMix~\cite{kim2020puzzle} and SaliencyMix~\cite{uddin2020saliencymix}. It applies Mixup based on the saliency~\cite{simonyan2014deep} of tokens. Our implementation is based on the official code of SSMix\footnote{\url{https://github.com/clovaai/ssmix}}.

\end{itemize}

\section{Implementation Details}
For all the experiments, we set the learning rate of the encoder model as 1e-5, set the learning rate of the two-layer MLP classifier as 1e-3, and tried different batch sizes within 8, 16, and 32 to choose the best performance. For the hyper-parameters in Dirichlet distribution and Beta distribution, we set $\tau$ as 1.0 and set $\alpha$ as 0.75. The coefficient $\gamma$ of JSD loss was set to be 8, and the max number of training epochs was set to be 20. All these hyper-parameters are shared among the models. All the experiments are performed multiple times across different seeds on a single NVIDIA RTX 8000 GPU.

\end{document}